\documentclass[10pt,twocolumn,letterpaper]{article}

\usepackage[pagenumbers]{wacv} 

\usepackage{tabularx}
\usepackage{multirow}
\usepackage{booktabs,tabularx,threeparttable}
\usepackage{pifont} 
\newcommand{\cmark}{\ding{51}}

\definecolor{wacvblue}{rgb}{0.21,0.49,0.74}
\usepackage[pagebackref,breaklinks,colorlinks,allcolors=wacvblue]{hyperref}

\def\confName{WACV}
\def\confYear{2026}

\begin{document}

\title{OpenLVLM-MIA: A Controlled Benchmark Revealing the Limits of Membership Inference Attacks on Large Vision-Language Models}

\author{Ryoto Miyamoto\\
Waseda University\\
Tokyo, Japan\\
{\tt\small r-miyamoto@toki.waseda.jp}
\and
Xin Fan\\
Waseda University\\
Tokyo, Japan
\and
Fuyuko Kido\\
Waseda University\\
Tokyo, Japan
\and
\and
Tsuneo Matsumoto\\
Hitotsubashi University\\
Tokyo, Japan
\and
Hayato Yamana\\
Waseda University\\
Tokyo, Japan
}

\maketitle

\begin{abstract}
OpenLVLM-MIA is a new benchmark that highlights fundamental challenges in evaluating membership inference attacks (MIA) against large vision-language models (LVLMs).
While prior work has reported high attack success rates, our analysis suggests that these results often arise from detecting distributional bias introduced during dataset construction rather than from identifying true membership status.
To address this issue, we introduce a controlled benchmark of 6{,}000 images where the distributions of member and non-member samples are carefully balanced, and ground-truth membership labels are provided across three distinct training stages.
Experiments using OpenLVLM-MIA demonstrated that the performance of state-of-the-art MIA methods approached chance-level.
OpenLVLM-MIA, designed to be transparent and unbiased benchmark, clarifies certain limitations of MIA research on LVLMs and provides a solid foundation for developing stronger privacy-preserving techniques.
\end{abstract}

\section{Introduction}
\label{sec:intro}

Recent progress in Large Language Models has paved the way for Large Vision-Language Models (LVLMs) that jointly process images and text. Commercial models such as Gemini~\cite{team2023gemini,team2024gemini} and the GPT family~\cite{openai2024gpt,openai2023gpt}, as well as the open-source model LLaVA~\cite{liu2023visual,liu2024improved}, achieve strong performance on many tasks, including captioning and visual question answering. These models are trained on hundreds of millions of images.

\subsection{Escalating Privacy Risks}

Training LVLMs requires large amounts of image data; thus, large-scale datasets such as LAION-5B~\cite{schuhmann2022laion} and Conceptual Captions~\cite{sharma2018conceptual} are used for training.
Since these datasets are collected via web crawling, there is a risk that confidential images, such as medical images, personal photos, and copyrighted works, may be unintentionally included in the training data.
Private data like patient diagnostic images and passport photos, as well as intellectual property, such as artworks and commercial photographs, may be used without the owners' consent.

A critical issue is that many vision-language models, including OpenAI CLIP~\cite{radford2021learning}, do not disclose the details of their training data. This lack of transparency prevents individuals and organizations from checking whether their images were used. Membership inference attacks (MIA) have therefore become an indispensable tool to quantify these privacy risks.

\subsection{Limitations of Existing MIA Research}

In recent years, several MIA techniques for LLMs~\cite{shi2023detecting,fu2025mia,fan2025dad} have been proposed, reporting high success rates. For LVLMs as well, Li et al.~\cite{li2024membership} achieved a maximum AUROC of 0.82 on their VL-MIA benchmark, which suggested that LVLMs present substantial privacy risks.
However, our analysis suggests that these high performance results may not necessarily reflect true membership detection. 

Our in-depth examination of existing benchmarks reveals the following two issues:

\textbf{1) Presence of distribution bias:} Member and non-member data were collected in different time periods and from different sources, introducing spurious attributes that make them separable for reasons unrelated to membership and thus undermine MIA evaluation.

\textbf{2) Uncertain ground truth:} Because many LVLMs keep parts of their training data undisclosed, the true membership of test data cannot be verified, making evaluation based on such benchmarks potentially ambiguous.

\subsection{Motivation and Goal}

The above two observations raise a research question: \emph{What do current MIA methods actually measure?} Does a high attack success rate truly detect model memory, or is it merely capturing biases in the data during dataset construction? To answer the research question, we construct OpenLVLM-MIA, a new controlled benchmark consisting of a benchmark dataset and a target model.

\subsection{Contributions}

We identify key challenges in LVLM MIA research and outline three contributions:

\textbf{1) OpenLVLM-MIA benchmark dataset:}
We introduced a controlled dataset comprising 6{,}000 images characterized by three defining properties: 
(i) the member and non-member subsets are distributionally matched; 
(ii) each example is annotated with a ground-truth membership label; and 
(iii) the dataset supports independent evaluation at multiple training stages. 
This design reduces distributional bias, enabling fair evaluation of membership inference attacks (MIAs).

\textbf{2) Limits of existing methods:}
Using our benchmark, comprehensive experiments showed that state-of-the-art MIA methods achieved only chance-level performance (AUROC $\approx 0.5$) under properly controlled conditions. This suggests that previously reported gains may reflect dataset artifacts rather than true membership detection.

\textbf{3) Community resources:}
We have released the dataset, evaluation tools, trained models, and experimental code. This enables the reproduction and verification of our results and provides a foundation for future research.

\section{Related Work}
\label{sec:related}

\subsection{Membership Inference Attacks}

Membership inference attacks (MIAs) were first formalized by Shokri et al.~\cite{shokri2017membership}.
Subsequently, Yeom et al.~\cite{yeom2018privacy} developed an efficient method that directly utilizes model outputs even under conditions where constructing shadow models is difficult.

With the emergence of large language models (LLMs), approaches such as Min-K\% Prob~\cite{shi2023detecting} were introduced. However, Duan et al.~\cite{duan2024membership} noted that these methods may rely heavily on temporal shifts and distributional differences.

\subsection{MIA Benchmark Datasets for Vision--Language Models}

Dedicated MIA benchmarks for large-scale vision--language models (LVLMs) remain scarce, with only the following datasets publicly available.

\subsubsection{VL-MIA Dataset}
VL-MIA~\cite{li2024membership} is the first MIA benchmark for LVLMs, targeting three systems (LLaVA v1.5~\cite{liu2024improved}, MiniGPT-4~\cite{zhu2023minigpt}, and LLaMA-Adapter v2~\cite{gao2023llamaadapterv2}). It contains 600 member and 600 non-member samples split into two subsets:
\begin{itemize}
\item \textbf{img\_Flickr}: Consisting of 300 member images (natural images used for training) and 300 non-member images (natural images collected after the model was trained)
\item \textbf{img\_dalle}: Consisting of 300 member images (web-crawled training images) and 300 non-member images (DALL-E--generated)
\end{itemize}

Since the training data for the above three systems is not fully public, non-members must be drawn from temporally mismatched or distributionally different sources (e.g., DALL-E--generated images).
Consequently, MIA methods may detect temporal shifts or differences between generated and natural images rather than true membership.

\subsubsection{Other Datasets}
Hu et al.~\cite{hu2025membership} and Zhu et al.~\cite{zhu2025revisiting} used datasets for instruction tuning of LVLM MIA in their experiments. However, these datasets do not account for the pre-training stage, and the datasets themselves have not been made publicly available.

\subsection{Effects of Training Stages on MIAs}

LLaVA~\cite{liu2023visual,liu2024improved} is trained in three stages: vision encoder pretraining, projector training, and instruction tuning.
Caldarella et al.~\cite{caldarella2024phantom} highlighted the importance of knowledge acquired during pretraining; however, a method for assessing how each stage affects the retention of membership information has yet to be proposed.

\section{Preliminary}
\label{sec:problem}

This section provides the problem formulation of membership inference attacks (MIA) against large vision-language models (LVLMs) , followed by discussing the distribution bias issue that arises in their evaluation.

\subsection{Problem Formulation}

A gray-box setting is considered, where the adversary can query the model to access its output logits. 
Given an image $v$ and the model’s response $f_\theta(v)$, the goal of an attack $\mathcal{A}$ is to determine whether $v$ was included in the training set:

\begin{equation}
\mathcal{A}: (v, f_\theta(v)) \rightarrow \{0, 1\}, 
\end{equation}
where 1 denotes a member (in-training) sample and 0 denotes a non-member sample.

\subsection{Distribution Bias Problem}

MIA evaluation faces a challenge called dataset \textbf{distribution bias}, which arises when the member and non-member samples are drawn from different distributions.
In this paper, we use the term \emph{temporal shift} as distribution differences caused by the time at which data were collected, which can lead to differences in the appearance and characteristics of the images.

Examples include:
\begin{itemize}
\item \textbf{Temporal shift}: member and non-member data are collected at different time periods
\item \textbf{Source mismatch}: data are collected from different datasets or domains
\end{itemize}

In such cases, an MIA method may capture these distributional differences rather than true membership information. 
This undermines the validity and practicality of MIA research.

\section{Proposed Benchmark Model: OpenCLIP-LLaVA}
\label{sec:model}

In this study, we construct a Large Vision–Language Model (LVLM) using only publicly available data to ensure reproducibility and transparency for membership inference attack (MIA) research on LVLMs.

\subsection{Limitations of Prior Work and Mitigations in the Proposed Benchmark Model}

Previous MIA studies for LVLMs face the following fundamental issues:

\begin{enumerate}
\item \textbf{Presence of distribution bias}: Member and non-member data are collected at different times or from different sources, making them distinguishable by factors unrelated to membership.
\item \textbf{Uncertainty of the ground truth}: Existing LVLMs include components trained on undisclosed data; thus, the true membership of test data cannot be verified.
\end{enumerate}

We address these issues as follows:

\textbf{Addressing Issue 1: distribution alignment.}
We mitigate distribution bias by selecting member and non-member images from data sources in the same time period and domain. Specifically, we choose non-member images from web-crawled datasets with characteristics similar to the member images, or from the validation split of the same dataset.

\textbf{Addressing Issue 2: Guaranteeing the ground truth of the membership.}
We construct an OpenCLIP-LLaVA model trained on public data and record whether each image is used during training. This guarantees ground-truth membership labels for all 6{,}000 images.

\subsection{OpenCLIP-LLaVA}

The OpenCLIP-LLaVA model follows the LLaVA architecture while training all components exclusively on public data. To guarantee the ground truth of the membership, we cannot use existing LVLMs, because they include components trained in private data, such as OpenAI CLIP~\cite{radford2021learning}. Therefore, we build an LVLM using only resources with public training data. Distribution alignment is handled by selecting non-member images from the same time period and domain as the member images. We ensure scale and diversity by collecting 1{,}000 member and 1{,}000 non-member images for each training stage (6{,}000 in total). The configuration is as follows:

\begin{itemize}
  \item \textbf{Vision Encoder}: OpenCLIP ViT-B/32~\cite{ilharco_gabriel_2021_5143773}
  \item \textbf{LLM}: Vicuna-7B v1.5~\cite{zheng2023judging}
  \item \textbf{Architecture}: LLaVA-compatible~\cite{liu2023visual}
\end{itemize}

\subsubsection{Training Datasets}

Following Liu et al.~\cite{liu2023visual}, the model is trained in three stages:
\begin{itemize}
  \item \textbf{Vision Encoder Pretraining}: LAION-2B-en (about 2.32B image–text pairs)~\cite{schuhmann2022laion}
  \item \textbf{Projector Pretraining}: LLaVA-Pretrain (558K image–text pairs)~\cite{liu2023visual}
  \item \textbf{Instruction Tuning}: LLaVA-Instruct (665K conversation pairs)~\cite{liu2024improved}
\end{itemize}

See the supplementary materials for detailed hyperparameters and settings for each stage.

\subsubsection{Performance comparison on LVLM Benchmarks}

Table~\ref{tab:benchmark} shows the OpenCLIP-LLaVA scores on the standard LVLM benchmarks and compares them with existing models. These results confirm that OpenCLIP-LLaVA achieves a performance level comparable to existing LVLMs. Therefore, the findings from the MIA evaluation of this model provide insights applicable to general LVLMs.

\begin{table}[t]
  \centering
  \caption{Performance comparison of OpenCLIP-LLaVA with existing models on LVLM benchmarks}
  \label{tab:benchmark}
  \small
  \setlength{\tabcolsep}{3pt}
  \begin{tabularx}{\columnwidth}{@{}l*{4}{>{\centering\arraybackslash}X}@{}}
    \toprule
    Model & MME Bench$\uparrow$ & MMMU Bench$\uparrow$ & MMVet Bench$\uparrow$ & LLaVA-Bench$\uparrow$ \\ \midrule
    OpenCLIP-LLaVA & 1570.9 & 28.9 & 21.7 & 76.1 \\
    LLaVA v1.5     & 1510.7 & 35.7 & 32.9 & 61.8 \\
    LLaVA v1       & 1075.5 & 34.1 & 28.3 & 57.2 \\
    MiniGPT-4 v1   &  968.4 & 23.6 & 15.6 & 45.1 \\ \bottomrule
  \end{tabularx}
  \vspace{1mm}
  \begin{minipage}{\linewidth}
    \footnotesize\raggedright
    Scores are computed using the official evaluation protocol/scripts of each benchmark (higher is better, $\uparrow$).
  \end{minipage}

\end{table}

\subsection{Public Resources}
We ensure reproducibility by releasing the following resources. Note that the dataset and model weights are hosted on Hugging Face and can be accessed using the code provided in the GitHub repository.

\begin{itemize}
    \item \textbf{Evaluation scripts and implementation code} \\
    \url{https://github.com/yamanalab/openlvlm-mia}
    
    \item \textbf{The OpenLVLM-MIA dataset and OpenCLIP-LLaVA (7B) model weights} \\
    Accessible via the code provided in the repository above.
\end{itemize}

These resources enable researchers to reproduce our experiments and facilitate further research on membership inference in LVLMs.

\section{Proposed Benchmark Dataset: OpenLVLM-MIA}
\label{sec:evaluation-datasets}

Existing MIA benchmarks for LVLMs suffer from two fundamental problems: 1) the true memberships are uncertain, and 2) there exists a distribution bias between member and non-member sets. As a result, MIA methods may detect such distribution bias in benchmark rather than genuine membership signals.

We propose \textbf{OpenLVLM-MIA}, the first benchmark that systematically addresses the above issues. It comprises 6{,}000 images (1{,}000 member and 1{,}000 non-member images for each of the three training stages) targeting an OpenCLIP--LLaVA model with a transparent training pipeline. Since the training data are public, we can verify the membership of each image, and careful data curation reduces distribution bias. This dataset makes it possible to fairly evaluate the true performance of MIA methods.

\subsection{Dataset Design}

The proposed OpenLVLM-MIA benchmark dataset enables independent evaluation of membership across the entire training pipeline, i.e., vision encoder pretraining, projector pretraining, and instruction tuning, because LVLM training typically proceeds in multiple stages, each relying on a different dataset. The key features are as follows:
\begin{itemize}
\item \textbf{High transparency}: All training data are public, guaranteeing membership ground truth.
\item \textbf{Three-stage evaluation}: It covers vision encoder pretraining, projector pretraining, and instruction tuning.
\item \textbf{Scale}: It consists of 1{,}000 member and 1{,}000 non-member images per stage (6{,}000 total).
\end{itemize}

\subsubsection{Member Data Collection}
The goal of member data collection is to identify the images actually used in each training phase and to guarantee true membership.
We selected 1,000 member images from each training dataset (LAION-2B-en, LLaVA-Pretrain, LLaVA-Instruct).
Specifically, we created MD5 hashes for all images and built a fast-searchable index using DuckDB~\cite{duckdb-software} to reliably track each image's training usage.
This system guarantees membership for the 1,000 images sampled evenly from each phase.

\subsubsection{Non-member Data Collection}
Non-member data collection aims to preserve distributional consistency with the member data while strictly guaranteeing that the images are not used for training. For each training phase, we selected 1,000 non-member images with characteristics similar to the member images. In the vision encoder and projector pretraining phases, we extracted images from the COYO-700M~\cite{kakaobrain2022coyo-700m} dataset, which was collected from the web around the same time as the member images. In the instruction tuning phase, we chose images from the validation split of the LLaVA-Instruct dataset that were guaranteed to be unused during training. Across all phases, we performed MD5 hash–based deduplication to confirm there was no overlap with the training data, securing 1{,}000 non-member images for each phase.

\subsection{Dataset Comparison}

\begin{table}[t]
\centering
\caption{Comparison of VL-MIA and OpenLVLM-MIA datasets}
\label{tab:dataset-comparison}
\small
\renewcommand{\arraystretch}{1.25}
\setlength{\tabcolsep}{4pt}
\begin{threeparttable}
\begin{tabularx}{\linewidth}{@{}l
  >{\centering\arraybackslash}p{2.1cm}
  >{\centering\arraybackslash}p{2.6cm}@{}}
\toprule
\textbf{Aspect} & \textbf{VL-MIA} & \textbf{OpenLVLM-MIA} \\
\midrule
Membership GT verified & N/A & \cmark \\
Distribution bias       & High   & Low    \\
Evaluation stages       & Final only & Three (separate) \\
\bottomrule
\end{tabularx}
\begin{tablenotes}[flushleft]
\footnotesize
\item GT = ground truth.  
\item “Three (separate)” = pretrain / finetune / inference evaluated individually.
\end{tablenotes}
\end{threeparttable}
\end{table}

Table~\ref{tab:dataset-comparison} shows the key differences between two datasets. In VL-MIA, the model's training data is private, making membership labels uncertain so that large distribution bias is introduced by mixing temporally mismatched data and DALL-E--generated images. In contrast, OpenLVLM-MIA adopts models trained on public data; thus, membership can be verified. By selecting data from the same period and domain, distribution bias is minimized. Moreover, while VL-MIA evaluates only the final model, OpenLVLM-MIA evaluates all three stages independently, enabling analysis of where and how membership information is retained. These improvements enable fair assessment of whether MIA methods detect membership rather than distribution bias.

\section{Experimental Evaluation}
\label{sec:experiments}
This study conducts two experiments to confirm the reliability of MIA evaluation for LVLMs: 1) a distribution audit that clarifies issues in existing datasets and highlights the advantages of our proposed dataset, and 2) a benchmark evaluation under a bias-controlled setting to measure true MIA performance. In Experiment~1, we use only visual features to quantify distribution bias in the VL-MIA dataset and to verify the distributional alignment of our OpenLVLM-MIA dataset. In Experiment~2, we systematically evaluate recent MIA methods under a bias-controlled setting and show that their performance is close to chance level.

\subsection{Experiment 1: Distribution Audit}

Verifying distribution bias between members and non-members is essential as a prerequisite for MIA evaluation.
If distribution bias exists, MIA methods may detect distribution differences rather than true membership.
We quantitatively verified this issue through visual feature-based evaluation using only image embeddings.

\subsubsection{C2ST Protocol with Visual Features Only}

We measure member/non-member separability using visual features alone, without any language features or model outputs.
We extract image embeddings from DINOv2-base~\cite{oquab2023dinov2}.
DINOv2 provides a text-independent, purely visual representation, thereby avoiding text-side bias.
Its self-supervised features capture generalizable representations of image features from low to high levels. Its high linear separability makes it well-suited for detecting distributional differences. 
This allows direct verification of whether “distinction is possible using images alone.”
After applying L2 normalization to each embedding vector, we evaluated the distinguishability between the two groups using the Classifier Two-Sample Test (C2ST)~\cite{lopez2017revisiting}.

Specifically, we use stratified 5-fold cross-validation with L2-regularized logistic regression ($C{=}1.0$) and prevent leakage via out-of-fold predictions.
We report AUROC, pAUROC@0.05~\cite{mcclish1989pauc}, and TPR@0.05FPR as the evaluation metrics, and evaluated the $p$-value at the $0.05$ level using a replacement test (1{,}000 times).

\subsubsection{Quantification of Distribution Differences}

In addition to C2ST, we quantify distribution differences using the following statistics:

\paragraph{Maximum Mean Discrepancy (MMD)~\cite{gretton2012kernel}}
MMD measures distance between distributions in the Reproducing kernel Hilbert space.
We use the RBF kernel $k(x, y) = \exp(-\gamma \|x-y\|^2)$ and set the kernel width $\gamma^{-1}$ by the median heuristic (the median of all pairwise distances).
Significance is evaluated via permutation tests at the 0.05 level.
\paragraph{Fréchet Distance (FID)~\cite{heusel2017gans}} FID is widely used for generative models. In this study, we use it to measure distances between distributions.
The embeddings of the member group and non-member group were approximated by multivariate normal distributions $\mathcal{N}(\mu_0, \Sigma_0)$ and $\mathcal{N}(\mu_1, \Sigma_1)$, respectively,
and the Fréchet distance was calculated using the following formula:
\begin{equation}
\text{FID} = \|\mu_1 - \mu_0\|^2 + \text{Tr}(\Sigma_0 + \Sigma_1 - 2(\Sigma_0\Sigma_1)^{1/2})
\label{eq:fid}
\end{equation}

A larger value indicates greater divergence between the distributions of the two groups.

\subsubsection{Results}

\paragraph{Quantitative Results}
Table~\ref{tab:distribution_audit} shows the distribution bias evaluation using DINOv2 embeddings.

\begin{table*}[t]
\centering
\caption{Distribution-bias evaluation using DINOv2 visual embeddings. Lower values indicate less unintended member/non-member separability (i.e., better distribution alignment).}
\label{tab:distribution_audit}
\small
\begin{tabular}{lcccc}
\toprule
\multirow{2}{*}{Dataset} & \multicolumn{2}{c}{C2ST Performance} & \multicolumn{2}{c}{Distribution Metrics} \\
\cmidrule(lr){2-3} \cmidrule(lr){4-5}
& AUROC$\downarrow$ & TPR@0.05FPR$\downarrow$ & MMD$\downarrow$ & FID$\downarrow$ \\
\midrule
VL-MIA: img\_Dalle & 0.729 & 0.240 & 0.0061 & 0.157 \\
VL-MIA: img\_Flickr & 0.949 & 0.813 & 0.0188 & 0.428 \\
\midrule
OpenLVLM-MIA: Vision Encoder & 0.554 & 0.063 & 0.0018 & 0.142 \\
OpenLVLM-MIA: Projector & 0.583 & 0.095 & 0.0022 & 0.152 \\
OpenLVLM-MIA: Instruction Tuning & 0.515 & 0.053 & 0.0016 & 0.137 \\
\bottomrule
\end{tabular}
\end{table*}

\paragraph{Bias in VL-MIA}
Large distributional bias was detected in the VL-MIA dataset.
For \texttt{img\_Flickr} (natural images), it achieved an AUROC of 0.949, enabling the identification of member/non-member with an accuracy of 0.813 (at 0.05 FPR) using only image embeddings.
For \texttt{img\_Dalle} (AI-generated images), AUROC=0.729 and TPR@0.05FPR=0.240.
The difference is significant at $p<0.05$.
MMD and FID are also large; in particular, \texttt{img\_Flickr} has FID=0.428, indicating distinct distributions.
These results suggest that reported MIA performance on VL-MIA may capture dataset bias rather than true membership.

\paragraph{Distributional Alignment in OpenLVLM-MIA}
In contrast, OpenLVLM-MIA shows fair distributional alignment.
Across all phases, TPR@0.05FPR stays between 0.053 and 0.095, indicating that the two groups are hard to distinguish even at low FPR. AUROC also remains around 0.5, confirming high distribution consistency.
MMD is at most 0.0022, showing effective bias control.
Differences are significant at $p<0.05$ for the Vision Encoder and Projector phases, but not for the final Instruction Tuning phase ($p>0.05$).
This supports fair MIA evaluation enabled by public training data and careful dataset design.

\begin{figure*}[t]
\centering
\includegraphics[width=\textwidth]{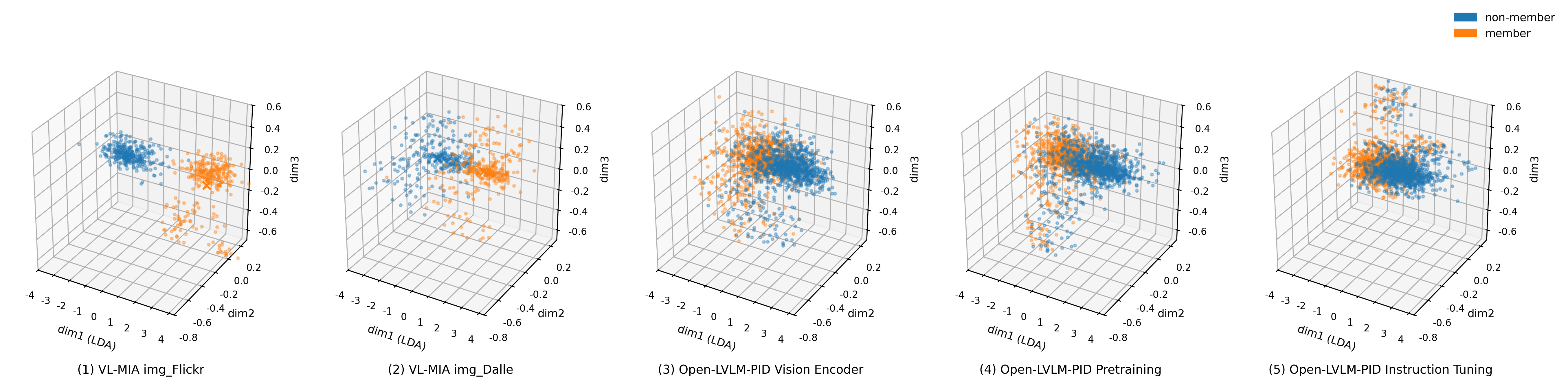}
\caption{LDA+PCA 3D visualization of the embedding space. \textit{dim1} is the supervised LDA discriminant axis; \textit{dim2--3} are the top two PCA components in the subspace orthogonal to the LDA axis. VL-MIA shows noticeable separation, while OpenLVLM-MIA exhibits considerable overlap.}
\label{fig:lda_pca_visualization}
\end{figure*}

\paragraph{Embedding-Space Visualization}
Figure~\ref{fig:lda_pca_visualization} visualizes the embedding space in 3D.
Nonlinear dimension reduction methods such as t-SNE and UMAP emphasize local structure, but image embeddings are high-dimensional, making it difficult to visualize separation in 3D space using unsupervised methods.
Therefore, in this study, we quantified separability using supervised dimensionality reduction with membership labels, where the Fisher discriminant axis $w \propto S_w^{-1}(\mu_1 - \mu_0)$ derived from LDA~\cite{fisher1936use} serves as the first axis.
Furthermore, by applying PCA~\cite{pearson1901liii} in the residual space orthogonal to this, we constructed the second and third axes, thereby visualizing variations beyond the discriminant axis.
This makes any separation (or its absence) visible.
The plots visually support the findings: VL-MIA displays a noticeable separation between member and non-member samples, whereas OpenLVLM-MIA exhibits considerable overlap.

\subsection{Experiment 2: Benchmark Evaluation}
In Experiment 2, we evaluated the accuracy of existing MIA methods on the proposed benchmark (OpenLVLM-MIA). The experiments revealed that existing methods achieved chance-level AUROC, suggesting they may have detected distribution bias on previous datasets.

\subsubsection{Setup}

\paragraph{Problem Setting}
The goal of MIA is to decide whether a given data point (image or text) was in the model's training set.
We assume a gray-box setting where the attacker can query the model and access output logits and generated text.
We compute membership scores from logit slices corresponding to the image embedding (\texttt{img}), instruction prompt to encourage captioning (\texttt{inst}), generated description (\texttt{desp}), and their concatenation (\texttt{inst+desp}).

\paragraph{Methods}
We evaluate ten state-of-the-art MIA methods on LVLMs:
Perplexity  (generative text confusion); Min-K\% Probability~\cite{shi2023detecting} (focusing on the lowest-probability tokens; $K{=}0,10,20$); and Max Rényi~\cite{li2024membership} (Rényi entropy-based; $\alpha{=}0.5,1.0$ with $K{=}0,10,100$).
We report AUROC~\cite{fawcett2006introduction} and TPR@0.05FPR~\cite{hand2009roc}.

\subsubsection{Results}
Table~\ref{tab:main_results} shows MIA performance on OpenLVLM-MIA.
Under the bias-controlled setting, we evaluate ten methods across three training phases (Vision Encoder, Projector, Instruction Tuning) and four logit slices (\texttt{img}, \texttt{inst}, \texttt{desp}, \texttt{inst+desp}).
Across all conditions, AUROC lies in the narrow range 0.407--0.527, at best only slightly above chance (0.5).

\begin{table*}[t]
\centering
\caption{MIA performance (AUROC) on OpenLVLM-MIA. Bold: highest; underlined: lowest per column.}
\label{tab:main_results}
\small
\setlength{\tabcolsep}{5pt}
\begin{tabular}{lcccccccccccc}
\toprule
& \multicolumn{4}{c}{Vision Encoder} & \multicolumn{4}{c}{Projector} & \multicolumn{4}{c}{Instruction Tuning} \\
\cmidrule(lr){2-5} \cmidrule(lr){6-9} \cmidrule(lr){10-13}
Method & img & inst & desp & inst+desp & img & inst & desp & inst+desp & img & inst & desp & inst+desp \\
\midrule
Perplexity & - & 0.474 & \underline{0.491} & 0.473 & - & 0.409 & 0.497 & 0.436 & - & \underline{0.496} & 0.503 & 0.506 \\
Min-0\% & - & \underline{0.455} & 0.506 & \underline{0.450} & - & \underline{0.407} & \underline{0.475} & \underline{0.411} & - & 0.511 & \underline{0.484} & 0.510 \\
Min-10\% & - & \underline{0.455} & 0.498 & 0.459 & - & \underline{0.407} & 0.482 & 0.417 & - & 0.511 & 0.493 & \underline{0.496} \\
Min-20\% & - & 0.461 & 0.494 & 0.465 & - & 0.414 & 0.486 & 0.422 & - & 0.504 & 0.495 & 0.502 \\
Max Rényi 0\% ($\alpha{=}0.5$) & 0.511 & \textbf{0.521} & 0.520 & 0.522 & \textbf{0.486} & \textbf{0.510} & 0.485 & \textbf{0.513} & \underline{0.488} & \textbf{0.513} & \textbf{0.527} & 0.521 \\
Max Rényi 10\% ($\alpha{=}0.5$) & 0.507 & \textbf{0.521} & 0.514 & \textbf{0.526} & 0.479 & \textbf{0.510} & 0.489 & 0.498 & 0.493 & \textbf{0.513} & 0.526 & 0.522 \\
Max Rényi 100\% ($\alpha{=}0.5$) & \underline{0.503} & 0.493 & 0.495 & 0.495 & \underline{0.462} & 0.476 & 0.481 & 0.474 & 0.510 & 0.508 & 0.518 & 0.517 \\
Max Rényi 0\% ($\alpha{=}1.0$) & 0.506 & 0.509 & \textbf{0.522} & 0.524 & 0.482 & 0.500 & 0.501 & 0.507 & 0.492 & 0.499 & 0.523 & \textbf{0.523} \\
Max Rényi 10\% ($\alpha{=}1.0$) & 0.509 & 0.509 & 0.516 & 0.519 & 0.483 & 0.500 & \textbf{0.503} & 0.506 & 0.499 & 0.499 & 0.515 & 0.517 \\
Max Rényi 100\% ($\alpha{=}1.0$) & \textbf{0.517} & 0.512 & 0.504 & 0.508 & 0.481 & 0.490 & 0.500 & 0.498 & \textbf{0.518} & 0.516 & 0.509 & 0.510 \\
\bottomrule
\end{tabular}
\end{table*}

In our carefully controlled benchmark, all state-of-the-art MIA methods converge to AUROC 0.41--0.52,  demonstrating performance indistinguishable from random guessing (Table~\ref{tab:main_results}).
The best case is 0.527 (Max Rényi 0\%, $\alpha{=}0.5$), while the worst is 0.407 (Min-0\%, Projector, \texttt{inst}).
These results suggest that, in this setting, membership inference is effectively uninformative under our controlled benchmark.

\begin{table*}[t]
\centering
\caption{MIA performance (TPR@0.05FPR) on OpenLVLM-MIA. Bold: highest; underlined: lowest per column.}
\label{tab:main_results_tpr}
\small
\setlength{\tabcolsep}{5pt}
\begin{tabular}{lcccccccccccc}
\toprule
& \multicolumn{4}{c}{Vision Encoder} & \multicolumn{4}{c}{Projector} & \multicolumn{4}{c}{Instruction Tuning} \\
\cmidrule(lr){2-5} \cmidrule(lr){6-9} \cmidrule(lr){10-13}
Method & img & inst & desp & inst+desp & img & inst & desp & inst+desp & img & inst & desp & inst+desp \\
\midrule
Perplexity & - & \underline{0.039} & \underline{0.040} & 0.043 & - & 0.026 & \underline{0.047} & 0.020 & - & \textbf{0.065} & 0.047 & 0.059 \\
Min-0\% & - & 0.045 & 0.054 & 0.039 & - & 0.026 & 0.050 & \underline{0.019} & - & 0.059 & 0.050 & 0.066 \\
Min-10\% & - & 0.045 & 0.048 & \underline{0.037} & - & 0.026 & 0.057 & 0.023 & - & 0.059 & 0.055 & 0.058 \\
Min-20\% & - & 0.045 & 0.045 & 0.042 & - & \underline{0.023} & \textbf{0.059} & 0.041 & - & 0.064 & 0.061 & 0.060 \\
Max Rényi 0\% ($\alpha{=}0.5$) & 0.059 & \textbf{0.078} & 0.043 & \textbf{0.063} & 0.041 & \textbf{0.075} & 0.031 & \textbf{0.053} & \underline{0.040} & 0.033 & 0.062 & \underline{0.041} \\
Max Rényi 10\% ($\alpha{=}0.5$) & 0.061 & \textbf{0.078} & 0.047 & 0.048 & 0.030 & \textbf{0.075} & \underline{0.029} & 0.046 & 0.051 & 0.033 & 0.054 & 0.042 \\
Max Rényi 100\% ($\alpha{=}0.5$) & 0.054 & 0.056 & 0.042 & 0.038 & 0.032 & 0.065 & 0.041 & 0.034 & 0.052 & 0.044 & 0.053 & 0.044 \\
Max Rényi 0\% ($\alpha{=}1.0$) & \underline{0.044} & 0.069 & 0.051 & 0.045 & 0.042 & 0.054 & 0.042 & 0.034 & 0.043 & \underline{0.020} & \textbf{0.066} & \textbf{0.068} \\
Max Rényi 10\% ($\alpha{=}1.0$) & 0.058 & 0.069 & 0.053 & 0.049 & \underline{0.025} & 0.054 & 0.043 & 0.047 & \textbf{0.057} & \underline{0.020} & 0.053 & 0.053 \\
Max Rényi 100\% ($\alpha{=}1.0$) & \textbf{0.061} & 0.066 & \textbf{0.065} & 0.057 & \textbf{0.051} & 0.057 & 0.040 & 0.045 & 0.049 & 0.045 & 0.052 & 0.048 \\
\bottomrule
\end{tabular}
\end{table*}

Table~\ref{tab:main_results_tpr} reports TPR@0.05FPR for practical operating points.
The best value is 0.078 (Max Rényi, Vision Encoder/Projector, \texttt{inst}), meaning that even at 95\% specificity most member samples (92.2\%) are missed.
In the Projector phase, more than half of the conditions fall below 0.04, with a minimum of 0.019 (Min-0\%, \texttt{inst+desp}).
These results indicate that MIAs are essentially ineffective in practical scenarios.

\paragraph{Analysis by Training Phase}
In the Vision Encoder phase, AUROC ranges from 0.450 to 0.526; Max Rényi methods are relatively higher but remain around 0.52.
The Projector phase shows the lowest performance, with AUROC 0.407--0.513 (minimum 0.407 for Min-0\% on \texttt{inst}); TPR@0.05FPR is below 0.04 in over half the conditions.
In the Instruction Tuning phase, AUROC narrows to 0.484--0.527 and differences between methods are minimal.

\section{Discussion}
\label{sec:discussion}

\subsection{Interpretation of Results}

\subsubsection{Effect of Distribution Bias}

One of the most significant findings of this study is the presence of severe distribution bias in existing VL-MIA datasets. The evaluation using only image embeddings achieves an AUROC of 0.949 (\texttt{img\_Flickr}), indicating that member/non-member can be identified with high accuracy without using any model outputs.
This result strongly suggests that the MIA performance reported in existing work may have captured systematic distribution bias stemming from differences in data collection methods or sources, rather than genuine membership inference.

In contrast, the evaluations on OpenLVLM-MIA showed C2ST AUROC values ranging from 0.515 to 0.583, with a value of 0.515 observed specifically during the Instruction Tuning phase.
This result shows that distribution bias can be effectively controlled through our appropriate dataset design.

\subsubsection{Difficulty of MIA for LVLMs}

When distribution bias is reduced, all state-of-the-art MIA methods fall within a narrow AUROC range (0.407--0.527), indicating the intrinsic difficulty of membership inference for LVLMs.
Two factors likely contribute: 1) a large volume of training data (billions of image--text pairs) dilutes the influence of any individual sample; and 2) cross-modal integration between vision and language can transform single-modality patterns in complex ways.

\subsection{Methodological Issues in MIA Evaluation}

Auditing the distribution of the dataset used to verify MIA is critically important due to the nature of the task.

Our visual feature-based distribution assessment is crucial for validating the assumptions behind MIA evaluation.
Prior studies implicitly assumed no distribution bias between member and non-member data, but our analysis shows this assumption often fails.
Future MIA studies should adopt the following protocol as a standard:

1) Collect member and non-member data from the close sources and time period.

2) Require a C2ST pretest using only visual features to verify minimal distribution bias. 

3) Reconsider and redesign the dataset if the C2ST AUROC is high.

\subsection{Analysis by Training Phase}

From the results, we obtain the following insights on how training phases affect MIA performance:

\textbf{Vision-encoder stage:} We observe a relatively wide AUROC spread (0.450--0.526), with the largest differences across MIA methods.
This is consistent with Caldarella et al.~\cite{caldarella2024phantom}, which suggests that pretraining contributes substantially to knowledge acquisition and may leave residual membership signals.

\textbf{Projector stage:} This stage shows the lowest AUROC (0.407--0.513).
Although the dataset used here showed the largest distribution bias in the audit, this stage’s contribution to membership signals appears small.
The effects of multimodal integrated learning and the fact that the weight of the projector is small relative to the entire model, resulting in a low proportion of training data interfering with the model, can also be considered contributing factors.

\textbf{Instruction-tuning stage:} Differences between methods are minimal (AUROC 0.484--0.527), and all methods show chance-level performance, which indicates large-scale additional data learning may overwrite initial membership signals.

\subsection{Future Work}

Current MIA approaches for large vision-language models are primarily limited to pure applications designed for language models and do not account for multimodality. Future development requires methods that explicitly leverage the interaction between vision and language. For example, AMIA~\cite{zhang-etal-2025-amia} uses text-image correlation for defense. Inverting such mechanisms could enable stronger attacks by detecting sensitivity in specific image regions or susceptibility to prompt variations. Analyzing patterns in caption generation under varying instructions and the consistency between visual attributes and linguistic descriptions may enable more effective membership inference. In addition, establishing standardized benchmarks with minimal distributional bias, such as OpenLVLM-MIA, is essential.

\section{Limitations}
\label{sec:limitations}

This work focuses only on LLaVA-1.5 (7B). We did not evaluate larger models (13B, 70B) or other architectures (e.g., BLIP-2, Flamingo).

Our dataset is limited to large-scale web-crawled images and does not explicitly include domain-specific data or data labeled as sensitive such that leakage would cause real-world harm.

The membership inference attacks (MIA) we evaluate assume a gray-box setting using standard prompts, focusing on output probabilities rather than prompt engineering~\cite{liu2024improved}; white-box attacks are out of scope.

We also do not study changes in membership after continual learning or fine-tuning, or the impact of defenses such as differential privacy.

\section{Conclusion}
\label{sec:conclusion}

We revealed a fundamental issue in how membership inference attacks (MIA) are evaluated for large vision–language models (LVLMs). Using a classifier two-sample test (C2ST) based only on visual features, we showed that existing vision–language MIA datasets exhibit an extreme distribution bias (AUROC = 0.949). This finding strongly suggests that the high attack success rates reported in prior work reflect detection of distribution bias between datasets rather than true membership inference.

To address this, we build \textbf{OpenLVLM-MIA}, a distribution-aligned benchmark of 6{,}000 images. We strictly align the distributions between member and non-member sets and guarantee membership at each training stage (vision encoder, projector, and instruction-tuning). Re-evaluating under proper conditions, we demonstrated that all existing MIA methods converge to random guessing; in particular, TPR@0.05FPR is at most 7.8\%, indicating that practical MIA is nearly impossible.

The findings of this study provide important implications for MIA research on LVLM.
First, distribution auditing should be established as a standard evaluation protocol, and dataset redesign should be considered when the C2ST AUROC is large.
Second, current MIA methods are primarily developed for single modalities, necessitating the development of new methods that leverage the interaction between vision and language.
Third, using models with fully disclosed training data enables evaluations that guarantee the ground truth of membership.

We have released the OpenLVLM-MIA dataset, evaluation code, and trained models to ensure reproducibility and transparency. This distribution-aligned benchmark provides a basis for measuring true membership inference ability. As future work, we will develop MIA methods that leverage the multimodal nature of LVLMs by analyzing vision–language interaction patterns and memory mechanisms at each training stage.

\newpage

{
    \small
    \bibliographystyle{ieeenat_fullname}
    \bibliography{main}
}

\end{document}